\documentclass{article}
\usepackage{amssymb}

\usepackage[final]{corl_2025} 
\usepackage{amsmath}
\usepackage{amsfonts}       
\usepackage{microtype}      
\usepackage{bm}
\usepackage{graphicx}
\usepackage{amsfonts}
\usepackage{multicol}
\usepackage{multirow}
\usepackage{booktabs}
\usepackage{graphicx}
\usepackage{array}
\usepackage{bm}
\usepackage{xspace}
\usepackage{float}
\usepackage{wrapfig}
\makeatother
\usepackage{makecell}
\usepackage{nicefrac}
\usepackage{xfrac}
\usepackage{caption}
 \usepackage{enumitem}
\usepackage{adjustbox}
\usepackage{booktabs}
\usepackage{multirow}
\usepackage[ruled]{algorithm2e}
\usepackage{epsfig}
\usepackage{color}
\usepackage{wrapfig,lipsum}
\usepackage{balance}
\usepackage{caption}
\usepackage{multirow}
\usepackage{graphicx}
\usepackage[table,xcdraw]{xcolor}
\usepackage{tocloft}
\hypersetup{
    colorlinks=true,
    linkcolor=blue,
    filecolor=magenta,      
    urlcolor=cyan,
    pdftitle={paperid-208},
    pdfpagemode=FullScreen,
    }
\newlength\savewidth

\definecolor{visual_color}{RGB}{144, 202, 249}
\definecolor{tactile_color}{RGB}{123, 31, 129}
\definecolor{tblblue}{RGB}{31, 119, 180}

\title{FlexiTac: A Low-Cost, Open-Source, Scalable Tactile Sensing Solution for Robotic Systems}
\vspace{-30pt}

\author{
Binghao Huang \quad \textbf{Yunzhu Li}  \\
\\
Columbia University 
}

\begin{document}

\maketitle


\begin{abstract}
We present \textbf{FlexiTac}, a low-cost, open-source, and scalable piezoresistive tactile sensing solution designed for robotic end-effectors. FlexiTac is a practical “plug-in” module consisting of (i) thin, flexible tactile sensor pads that provide dense tactile signals and (ii) a compact multi-channel readout board that streams synchronized measurements for real-time control and large-scale data collection. FlexiTac pads adopt a sealed three-layer laminate stack (FPC–Velostat–FPC) with electrode patterns directly integrated into flexible printed circuits, substantially improving fabrication throughput and repeatability while maintaining mechanical compliance for deployment on both rigid and soft grippers. The readout electronics use widely available, low-cost components and stream tactile signals to a host computer at 100 Hz via serial communication. 
Across multiple configurations, including fingertip pads and larger tactile mats, FlexiTac can be mounted on diverse platforms without major mechanical redesign. We further show that FlexiTac supports modern tactile learning pipelines, including 3D visuo-tactile fusion for contact-aware decision making, cross-embodiment skill transfer, and real-to-sim-to-real fine-tuning with GPU-parallel tactile simulation. Our project page is available at \url{https://flexitac.github.io/}.
\end{abstract}

\begin{figure}[h]
    \centering
    \includegraphics[width=\linewidth]{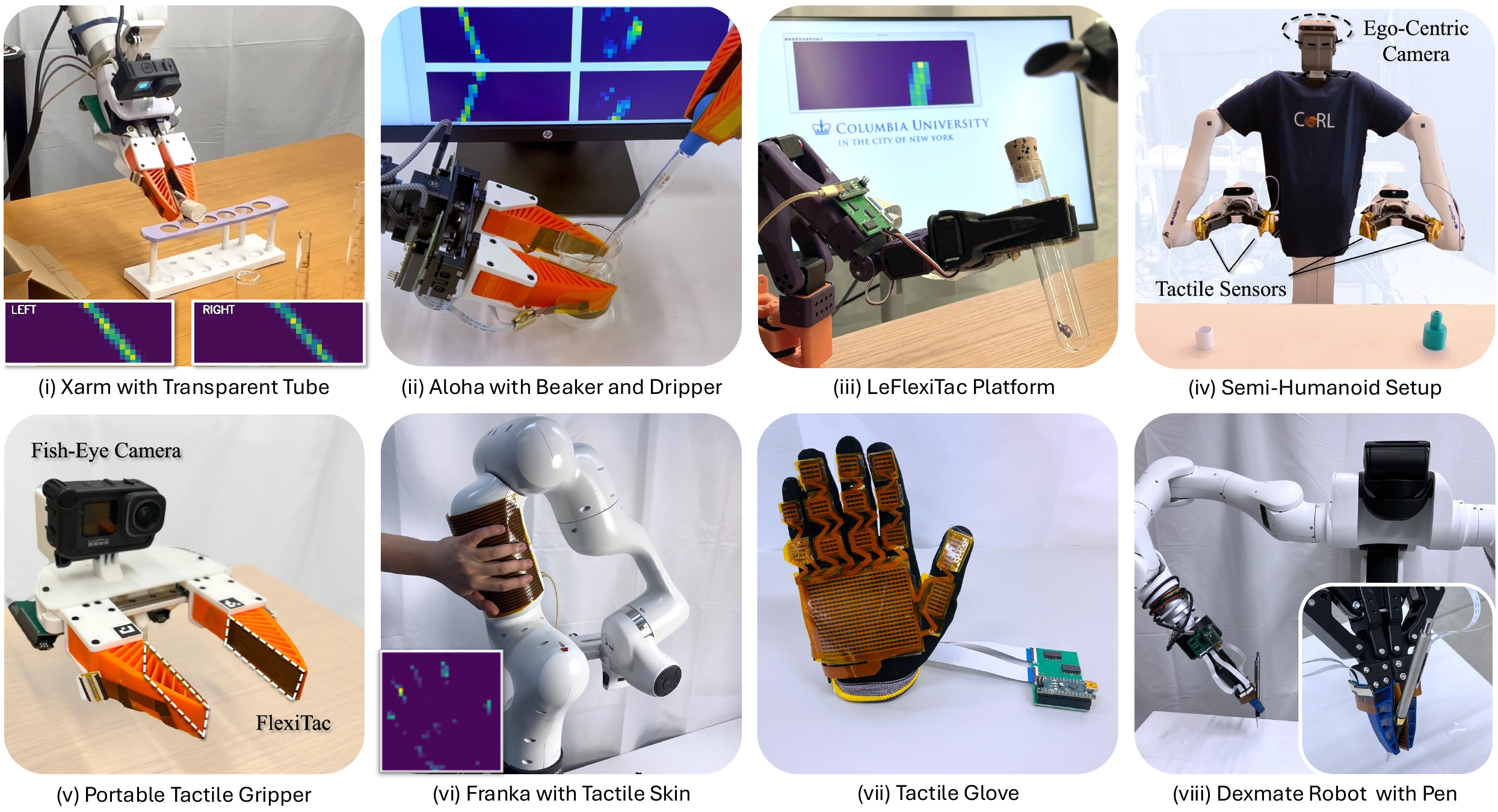}
    \vspace{-5pt}
\caption{\textbf{FlexiTac deployments across diverse platforms.}
We showcase FlexiTac integrated on multiple robot end-effectors and a wearable collector, spanning tabletop manipulation, bimanual coordination, mobile manipulation, and in-the-wild data acquisition. The breadth of deployments highlights the sensor’s conformability, modularity, and plug-and-play integration.}

    \label{fig:teaser}
    \vspace{-10pt}
\end{figure}

\clearpage

\vspace{-5pt}
\section{Introduction}
\label{sec:intro}
\vspace{-5pt}

Human skin provides rich, high-bandwidth feedback that enables dexterous behaviors such as gently grasping fragile objects, detecting incipient slip, and reorienting items in-hand without continuous visual monitoring. Such tactile feedback plays a central role in human grasp stability and force regulation, where visual and somatosensory cues jointly guide fingertip force control during object interaction~\cite{Jenmalm4486,Johansson2004RolesOG}. In contrast, despite rapid progress in learning-based manipulation and robotic hardware, today’s robot end-effectors still lag far behind human-level dexterity. Two trends have driven recent improvements: (i) progress in data-driven learning (e.g., imitation learning and RL fine-tuning) together with scalable data collection methods, and (ii) the increasing use of tactile sensing to complement vision, reduce occlusions, and enable closed-loop, contact-aware manipulation, providing the “last-millimeter” feedback needed for fine-grained precision.

Although robotic tactile sensors remain far less capable than biological skin, and typically capture only a subset of signals such as pressure rather than vibration, temperature, or texture, even partial tactile feedback has been shown to improve grasp outcome prediction, grasping and regrasping, contact-rich insertion, and visuo-tactile policy learning~\cite{calandra2017feeling,calandra2018learning,lee2019making,huang3d,sunil2023visuotactile,chen2025multi}. A wide variety of tactile sensing technologies have been explored, including 
vision-based tactile sensors~\cite{yuan2017gelsight,lambeta2020digit,ward2018tactip,ma2024gellink,padmanabha2020omnitact,taylor2022gelslim,buscher2015aug}, 
magnetic tactile sensors~\cite{bhirangi2021reskin,bhirangi2024anyskin,pattabiraman2025eflesh,tomo2018new,bhirangi2022all}, 
capacitive tactile arrays~\cite{wistreich2025dexskin}, and flexible piezoresistive tactile sensor arrays~\cite{luo2021learning,zlokapa2022integrated,Murphy2025fits}. 
Despite this progress, the field has not converged on a single dominant sensing modality. Among these options, flexible tactile sensors have become increasingly popular due to their compact form factor, ease of integration on existing grippers, simple fabrication process and mechanical structures, and robustness in contact-rich manipulation settings.

In this paper, we present FlexiTac, a low-cost, open-source, and scalable piezoresistive tactile sensing solution designed for robot end-effectors. As shown in Fig.~\ref{fig:teaser}, FlexiTac can be integrated into a wide range of robotic embodiments, from parallel-jaw grippers to multi-finger hands, highlighting the versatility of the platform across manipulation settings. FlexiTac consists of (1) a flexible sensor pad that provides dense tactile signals and (2) a compact, multi-channel readout board for synchronized data acquisition. Together, these components form a practical “drop-in” tactile module that supports both real-time control and large-scale data collection.

Our goal is to support research on complex contact-rich manipulation by making dense tactile sensing easier to deploy, reproduce, and scale across robot platforms. To this end, we focus on improving sensor robustness, simplifying manufacturing and assembly, and enabling simulation support. We further introduce multiple FlexiTac configurations, each optimized for different trade-offs among spatial resolution, durability, wiring complexity, and integration effort, and discuss their respective advantages and recommended use cases. The main contributions of this work are:

\textit{(i)} We present \textbf{FlexiTac}, an open-source and low-cost tactile sensing platform for robotics research that integrates flexible tactile sensor pads, compact readout electronics, and unified software interfaces for real-time, synchronized tactile acquisition. 

\textit{(ii)} We propose a manufacturing-friendly FPC-based design for piezoresistive tactile sensors that improves repeatability, scalability, and ease of customization across embodiments.

\textit{(iii)} We demonstrate that FlexiTac supports modern tactile learning pipelines, including visuo-tactile learning, cross-embodiment skill transfer, and real-sim-real pipeline with tactile simulation support.

\vspace{-5pt}
\section{Hardware}
\vspace{-5pt}
\label{sec:hardware}
\subsection{Sensor Principle and Pad Manufacturing}

\begin{figure}[t]
    \centering
    \vspace{-5pt}
    \includegraphics[width=\linewidth]{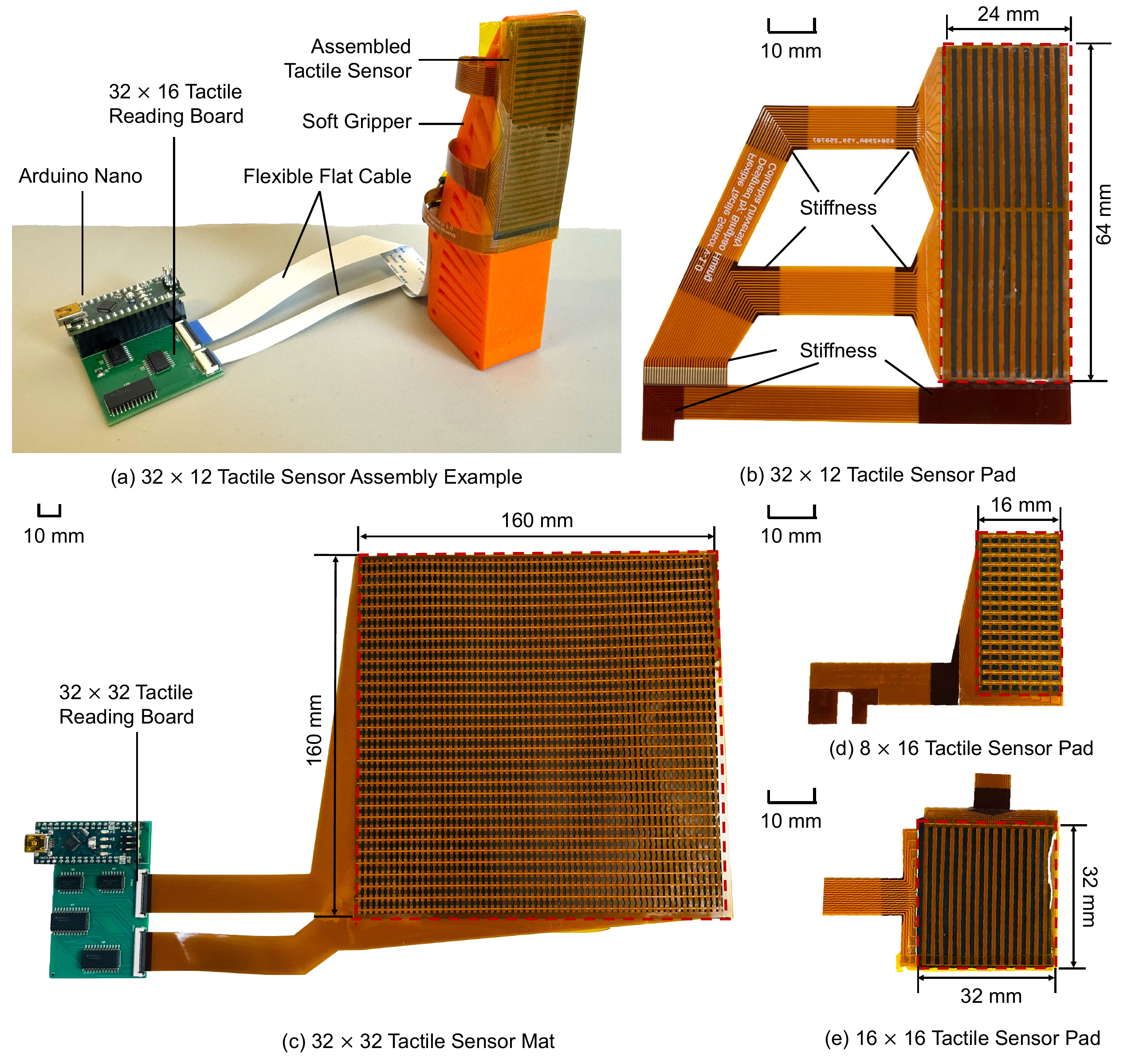} 
    \vspace{-15pt}
    \caption{\textbf{FlexiTac System Configurations.}
    \textit{(a)} Example assembly of a 32$\times$12 tactile pad integrated on a soft fin-shaped gripper, connected via a flexible flat cable (FFC) to a multi-channel readout board (with an Arduino Nano).  \textit{(b)} Close-up of the 32$\times$12 pad geometry.  \textit{(c)} A larger 32$\times$32 tactile mat configuration with its corresponding readout board.  \textit{(d-e)} Additional compact pad form factors (8$\times$16 and 16$\times$16) illustrating modular sizing options for different end-effectors.}
    
    \label{fig:sensor_assembly}
    \vspace{-10pt}
\end{figure}

FlexiTac sensor pads are based on a piezoresistive sensing matrix that converts applied mechanical
pressure into changes in electrical resistance; the readout electronics then convert these changes into
voltage signals. This sensing principle follows a broader line of scalable piezoresistive tactile arrays, where force-sensitive films are addressed through orthogonal electrodes to obtain spatially organized pressure measurements~\cite{sundaram2019Learning, buscher2015aug}. The pads are designed to be thin (total thickness $<1\,\text{mm}$), lightweight, and conformable, enabling straightforward integration onto a wide range of robotic end-effectors. As illustrated in Fig.~\ref{fig:teaser}, we mount FlexiTac on several common manipulation platforms, including the Robotiq 2F-140, xArm grippers, the ALOHA system, and the LeRobot gripper, without requiring significant mechanical redesign. Beyond rigid grippers, the sensor pads can also be deployed on compliant and deformable interfaces. In the example shown in Fig.~\ref{fig:sensor_assembly}, we integrate FlexiTac onto a soft, fin-shaped gripper. This mechanical compliance, together with the thin form factor, makes FlexiTac versatile for tactile sensing across diverse robotic applications.

Each finger of the gripper is equipped with a sensor pad comprising a $12 \times 32$ sensing matrix. The pad geometry, sensing density, and spatial resolution are customizable; in our latest design (FlexiTac V2), the pitch is set to $2\,\mathrm{mm}$ between adjacent sensing units. As illustrated in Fig.~\ref{fig:sensor_pad}(a), FlexiTac pads follow a sealed three-layer stack-up: a piezoresistive film (Velostat) is sandwiched between two flexible printed circuits (FPCs), and the entire laminate is encapsulated using laminating sheets. This fabrication process is simple and fast, and can typically be completed within $\sim$5 minutes per pad. A key innovation in FlexiTac V2 is the use of FPCs with patterned copper openings that serve directly as electrodes. Compared to our earlier design that relied on manually aligned conductive threads~\cite{huang3d}, this electrode-integrated FPC substantially reduces alignment overhead and improves manufacturing throughput and repeatability. Details of the FPC layout and electrode design are provided in Sec.~\ref{sec:fpc_design}.

To further improve manufacturing scalability, we use a desktop cutting machine (Silhouette Cameo 5) to rapidly fabricate sensor pad components with consistent dimensions, including the piezoresistive film and laminating sheets. This tool-assisted workflow reduces manual labor and enhances fabrication uniformity across batches.

\subsection{Flexible Printed Circuit Design}
\label{sec:fpc_design}

\begin{figure}[t]
    \centering
    \vspace{-5pt}
    \includegraphics[width=\linewidth]{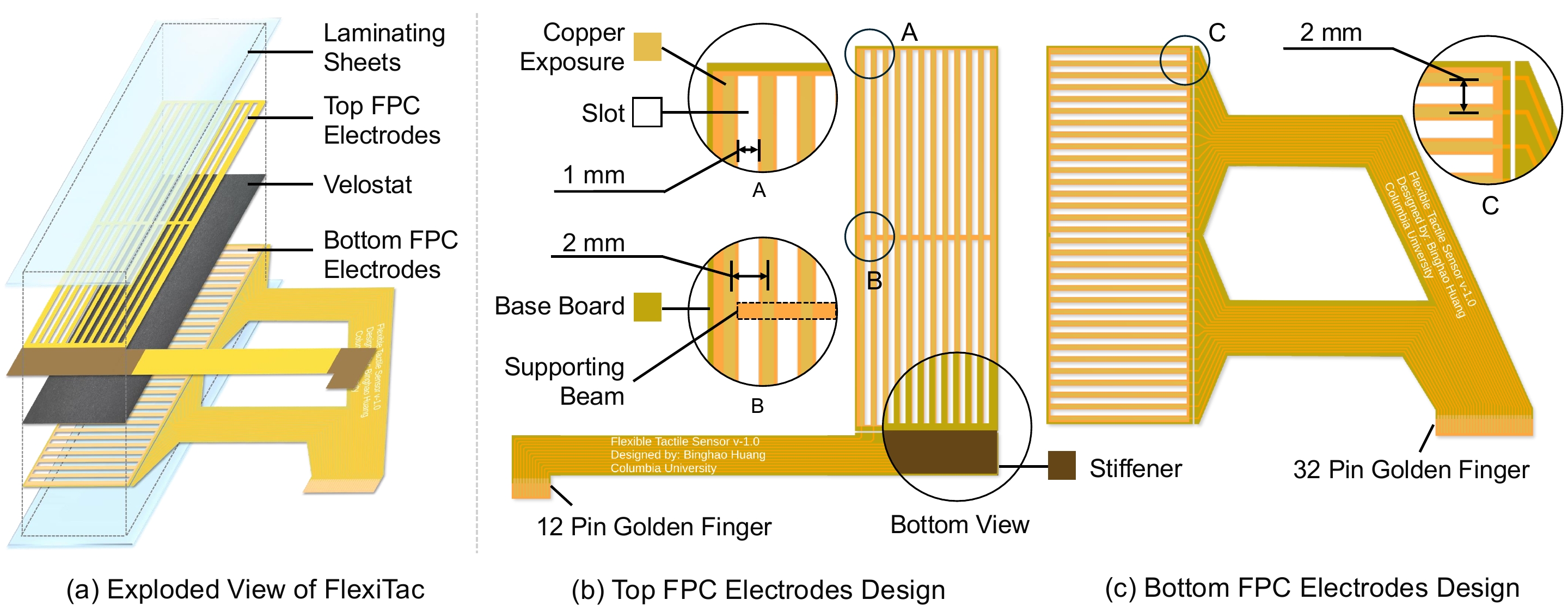} 
    \vspace{-15pt}
    \caption{\textbf{FlexiTac construction and electrode layout.}
    \textit{(a)} Exploded view of the sealed three-layer stack-up: top FPC electrodes, piezoresistive film (Velostat), and bottom FPC electrodes, encapsulated by laminating sheets, with stiffeners or golden fingers for reliable electrical interfacing. \textit{(b-c)} Top and bottom FPC electrode designs forming an orthogonal sensing matrix; each electrode intersection defines one taxel. Slots between adjacent electrodes improve conformability and concentrate deformation to increase sensitivity, while a supporting beam/stiffener improves structural reliability.}

    \label{fig:sensor_pad}
    \vspace{-10pt}
\end{figure}

Flexible printed circuits (FPCs) enable rapid, repeatable manufacturing and provide a convenient substrate for integrating dense electrode patterns. To accelerate sensor fabrication, FlexiTac V2 adopts FPCs with patterned copper openings that directly serve as electrodes, eliminating the need for manually aligned conductive threads. Fig.~\ref{fig:sensor_pad}(b,c) illustrates the electrode layout. The electrode pitch is $2\,\mathrm{mm}$, and two orthogonal FPC layers are used: on the top FPC, electrode traces extend vertically, while on the bottom FPC they extend horizontally. Each intersection between a top and bottom electrode defines one sensing unit in the tactile array. The pad contains 32 horizontal electrodes and 12 vertical electrodes, resulting in a $12 \times 32$ tactile array.

We further introduce narrow slots between adjacent electrodes. These slots serve two purposes: (i) they increase the compliance of the FPC layers, improving conformability on curved or soft surfaces, and (ii) they locally concentrate deformation onto the piezoresistive film (Velostat), which increases the effective pressure applied under contact and improves sensitivity.

Because the bottom FPC contains longer electrode traces, it is more compliant and can bend more easily. To maintain sufficient structural stiffness and prevent unwanted warping during assembly and use, we incorporate a supporting beam in the middle. In addition, we add 0.2 mm polyimide stiffeners on the bottom layer of the FPCs to increase the reliability of the tactile sensors. These reinforcements help preserve the intended pad geometry while keeping the active sensing area flexible.

\subsection{Readout Board PCB Design}

\begin{figure}[t]
    \centering
    \vspace{-5pt}
    \includegraphics[width=\linewidth]{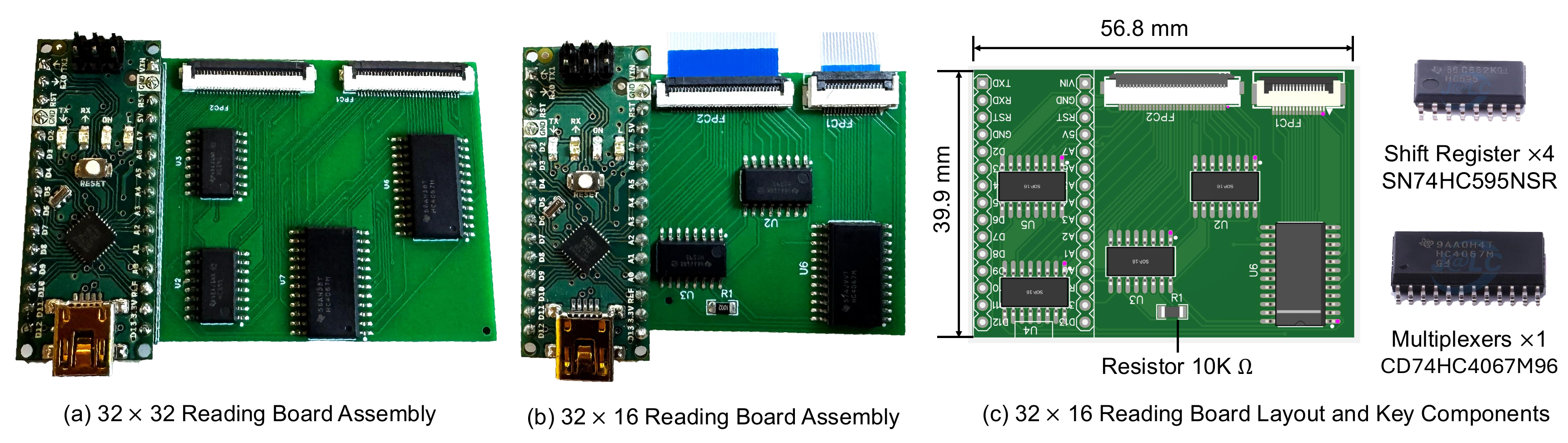} 
    \vspace{-15pt}
    \caption{\textbf{Multi-channel readout electronics for scalable tactile acquisition.}
\textit{(a)} 32$\times$32 readout board assembly. \textit{(b)} 32$\times$16 readout board assembly. \textit{(c)} Layout and key components of the 32$\times$16 board, including shift registers and a multiplexer for efficient addressing of high-dimensional tactile matrices with minimal wiring. The board streams synchronized tactile measurements to a host computer (100 Hz) via serial communication and is backward-compatible with lower-resolution pads by using a subset of channels.}

    \label{fig:reading_board}
    \vspace{-10pt}
\end{figure}

The readout board acquires the tactile signals and streams them to the host computer via serial communication at $100\,\mathrm{Hz}$. Sensor pads are connected to the board through standard 0.5 mm flexible flat cable (FFC) connectors, enabling reliable, low-profile multi-channel interfacing.
To lower the barrier to tactile sensing, we simplify not only the sensor pads but also the readout electronics. As shown in Fig.~\ref{fig:reading_board}, the overall circuit is intentionally lightweight and built from widely available, low-cost components. We use an Arduino Nano as the microcontroller (MCU), chosen for its broad adoption, strong community support, and ease of integration.

The readout board is primarily composed of multiplexers and shift registers, which together enable scalable addressing of a high-dimensional sensing matrix with minimal wiring and GPIO usage. Taking the $32 \times 16$ readout board  shown in Fig.~\ref{fig:reading_board}(c) as an example, this configuration uses one 16-channel multiplexer for analog signal routing and four 8-bit shift registers for digital addressing. Importantly, the same board design is backward-compatible with lower-resolution pads (i.e., any resolution up to $32 \times 16$) by configuring only a subset of channels. This minimalist design improves reproducibility and lowers the barrier to replication, modification, and extension by other researchers. It also enables a compact PCB footprint, making the readout board easier to mount on end-effectors or integrate into space-constrained robot hands.

\subsection{Cost Breakdown and Scalability}

To improve the scalability of FlexiTac, we focus not only on simplified fabrication and design, but also on keeping the system low-cost. A complete FlexiTac unit costs approximately \$30, with the main expenses coming from the sensor pad and the readout board. For the sensor pad, a pair of FPCs costs \$3.55 at a 30-unit order volume, and \$1.36 at a 1000-unit volume. For the readout board PCB, the cost is \$3.10 per board at 30 units and \$2.61 per board at 1000 units, excluding the microcontroller. Using an off-the-shelf Arduino Nano (approximately \$25), the total cost per tactile sensor is therefore around \$30. The readout cost could be further reduced by integrating the microcontroller and the remaining circuitry onto a single PCB, eliminating the separate Arduino module. Our current design intentionally keeps the Arduino as a modular component to balance cost with ease of assembly, debugging, and replication.


\vspace{-5pt}
\section{System-Level Integration for Robot Learning}
\label{sec:system}
\vspace{-5pt}

Beyond the sensor hardware itself, FlexiTac forms a practical tactile interface for robot learning systems. Its thin, flexible pads can be mounted on diverse end-effectors, while the readout electronics provide synchronized tactile streams that integrate naturally with vision, proprioception, and control pipelines. Rather than focusing on a single task, we highlight three representative capabilities enabled by FlexiTac: visuo-tactile representation learning, cross-embodiment skill transfer, and real-to-sim-to-real training with tactile simulation.

\subsection{Visuo-Tactile Representation and Decision Making}

Prior visuo-tactile learning methods have explored combining visual and tactile inputs for grasp prediction, regrasping, contact-rich manipulation, and in-hand manipulation~\cite{lin2024learning,yuan2023robot,falco2017cross}. As one example of visuo-tactile integration, FlexiTac can be mounted on fin-shaped grippers in a bimanual manipulation setup and combined with multi-view RGB-D cameras for large-scale data collection and learning-based control~\cite{huang3d}.
The system is designed as a modular multimodal stack: (i) tactile pads provide dense, contact-local signals on the gripper surface, (ii) the readout board streams synchronized tactile measurements to the host at a fixed rate, and (iii) the vision subsystem provides global scene geometry via multi-view depth. In practice, integration requires only lightweight attachment of the pads to the fingertips and a one-time calibration of the pad-to-gripper transform, after which each taxel can be associated with a 3D location on the end-effector over time.

\begin{figure}[t]
    \centering
    \vspace{-5pt}
    \includegraphics[width=\linewidth]{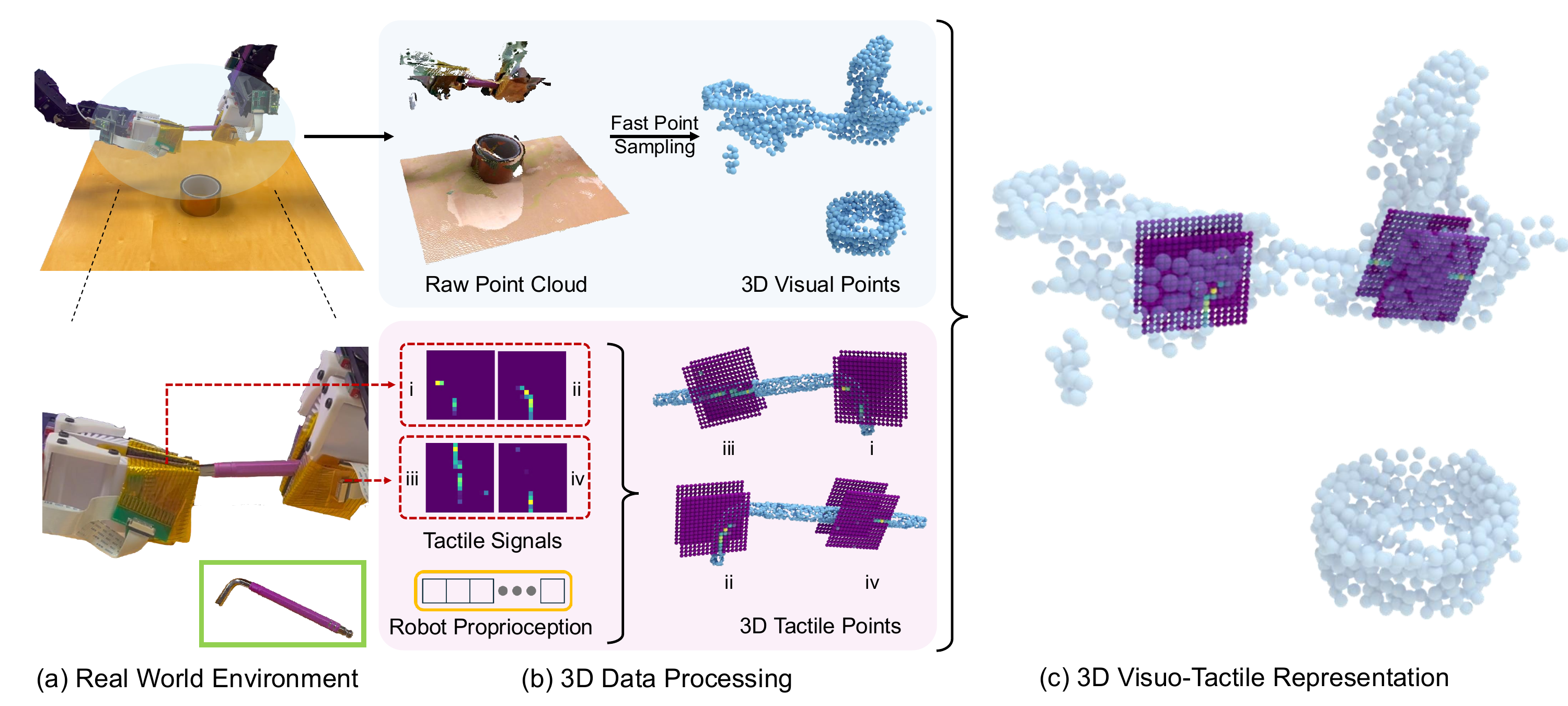} 
    \vspace{-10pt}
    \caption{\textbf{3D visuo-tactile fusion for contact-aware policies (3D-ViTac pipeline).}
\textit{(a)} Real-world environment with multimodal observations. \textit{(b)} Processing pipeline: multi-view RGB-D is reconstructed into a 3D visual point cloud, while tactile signals are transformed into 3D space using robot proprioception, attaching tactile signals as a feature. \textit{(c)} Unified 3D visuo-tactile point representation by merging visual and tactile points, enabling learning-based control.}

    \label{fig:3d-vitac}
    \vspace{-10pt}
\end{figure}

To enable principled multimodal fusion, we represent both vision and touch in a shared 3D space. As shown in Fig.~\ref{fig:3d-vitac}, visual observations are converted into a 3D point cloud reconstructed from multi-view depth. Tactile readings are lifted into 3D as a set of \emph{tactile points}: each taxel is assigned a 3D position via forward kinematics (and the calibrated pad geometry), while its tactile magnitude is appended as an additional feature channel. We then merge the visual point cloud and tactile point cloud into a unified 3D visuo-tactile representation, augmented with a modality indicator for each point, shown in Fig.~\ref{fig:3d-vitac}(c). This fusion is computationally lightweight and geometrically interpretable: each tactile unit corresponds to a spatial point carrying contact-intensity information, and the dense tactile array forms a contact-intensity distribution over the gripper surface, allowing the policy to reason explicitly about the spatial relationship between contact and scene geometry.

Given this fused 3D representation, action generation is performed by a learning-based decision module. In 3D-ViTac~\cite{huang3d}, we use a diffusion-policy formulation conditioned on the unified visuo-tactile point set. A point-cloud backbone encodes the fused observation and outputs a sequence of manipulation actions, enabling closed-loop corrections when vision is occluded or when precise contact regulation is required. Beyond task-specific imitation learning, recent work has explored broader paradigms for leveraging
tactile feedback, including self-supervised and contrastive pretraining for tactile or visuo-tactile
representations~\cite{lee2019making,guzey2023dexterity,dave2024multimodal,george2024visuo},
tactile integration into vision-language-action (VLA) models for contact-aware manipulation~\cite{zhang2026tacvla,tao2026leflexitac},
and reinforcement-learning-based visuo-tactile control~\cite{hoof2016stable,hansen2022visuo,chen2022visuo,dexpoint}.

\subsection{Cross-Embodiment Tactile System}

\begin{figure}[t]
    \centering
    \vspace{-5pt}
    \includegraphics[width=\linewidth]{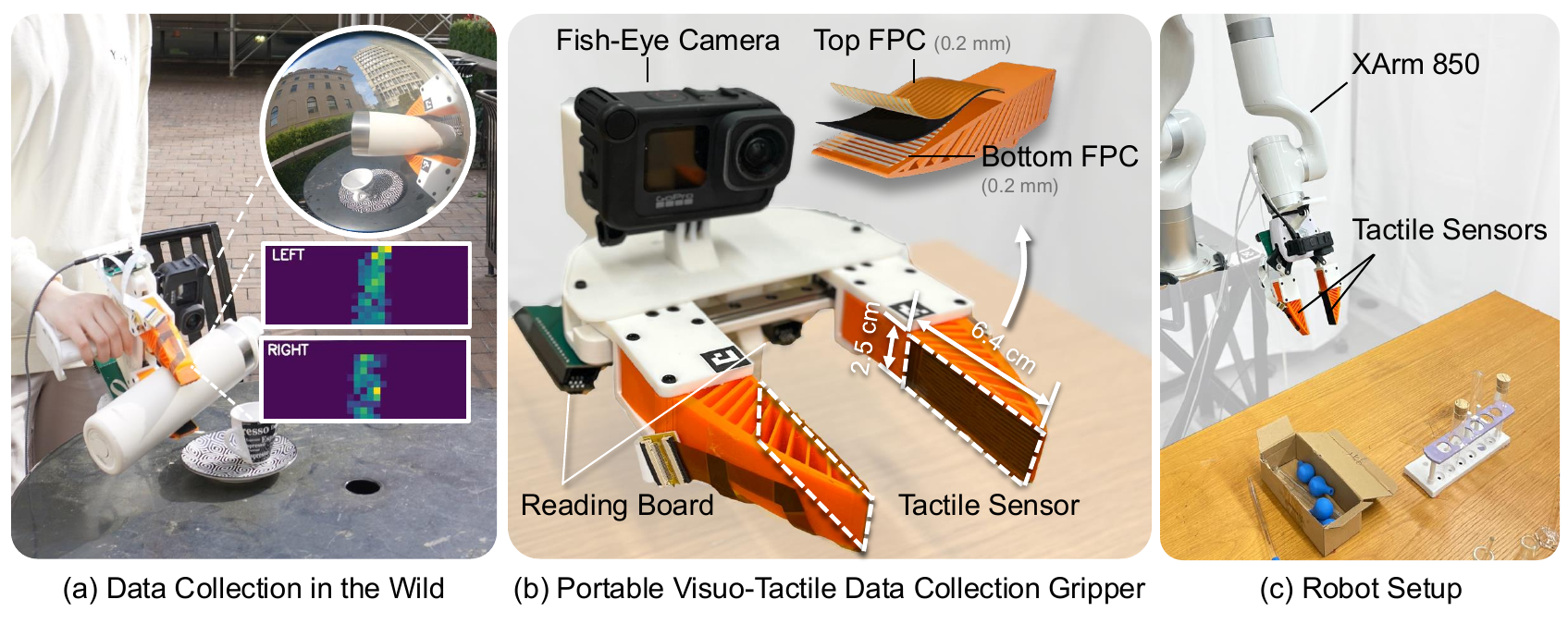} 
    \vspace{-15pt}
    \caption{\textbf{Cross-embodiment tactile sensing with a portable device and robot.} \textit{(a)} In-the-wild data collection using a portable visuo-tactile gripper. \textit{(b)} Portable device details showing a fisheye camera, FlexiTac pads, and a compact readout board for synchronized logging. \textit{(c)} Robot deployment on an xArm 850 equipped with FlexiTac. The shared sensing module supports consistent tactile signals across embodiments, enabling policy transfer from human demonstrations to robot execution.}
    \label{fig:umi-tactile}
    \vspace{-5pt}
\end{figure}

Recent cross-embodiment robot learning systems have explored scalable and portable data collection for transferring demonstrations across embodiments~\cite{chi2024universal,wang2024dexcap}. Similarly, FlexiTac can be deployed on both a portable human data-collection device and an
xArm-based robotic platform, providing a shared tactile interface across embodiments~\cite{zhu2025touch,kang2026learning}. We use (i) a portable human data-collection device that captures synchronized visual observations, actions, and dense fingertip tactile signals during natural manipulation (Fig.~\ref{fig:umi-tactile}(a,b)); and (ii) an xArm-based robotic platform equipped with a FlexiTac-instrumented gripper (Fig.~\ref{fig:umi-tactile}(c)). The same sensing stack is used on both platforms, allowing tactile observations to be recorded in a shared format and aligned with visual observations and control signals.

This shared tactile interface is important for transferring manipulation skills from human demonstrations to the robot. Successful transfer requires not only compatible visual observations and action representations across embodiments, but also tactile signals that remain reliable despite differences in sensor instances, mounting geometries, and data-collection sessions. In addition, the tactile hardware must withstand repeated contact and sliding during long-term use.

FlexiTac addresses these requirements through its dense array structure, which provides stable, spatially organized contact measurements over the fingertip or gripper surface. This representation enables the policy to learn contact-dependent manipulation skills from demonstrations
collected on a portable human-operated device, including grasp stabilization, pressure modulation, and
fine corrective adjustments. These skills can then be transferred to the robot through the shared tactile
interface.

\subsection{Tactile Simulation with Real-to-Sim-to-Real Pipeline}

\begin{figure}[t]
    \centering
    \vspace{-5pt}
    \includegraphics[width=\linewidth]{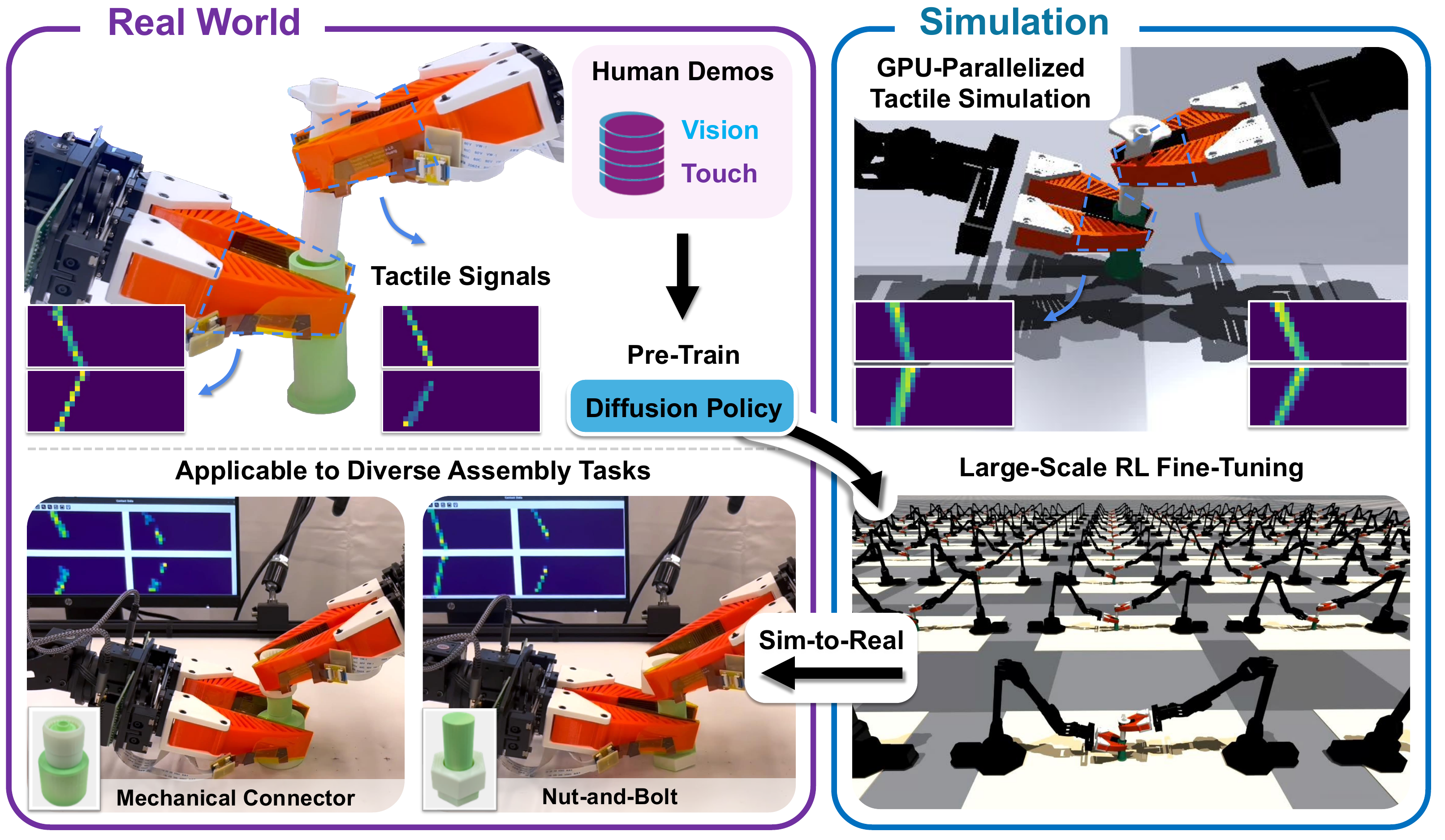} 
    \vspace{-15pt}
    \caption{\textbf{Real-to-sim-to-real learning pipeline with tactile simulation.}
Overview of a pipeline that combines real-world visuo-tactile demonstrations with GPU-parallel tactile simulation for scalable RL fine-tuning. Tactile signals are simulated at the taxel level to match the real sensor layout, enabling pre-training (e.g., diffusion policy), simulation-based fine-tuning on diverse assembly tasks (e.g., nut-and-bolt), and sim-to-real transfer aided by tactile calibration and consistent normalization between real and simulated signals.}

    \label{fig:vt-refine}
    \vspace{-10pt}
\end{figure}

Tactile simulation has been studied for both optical tactile sensors and contact-force-based tactile control, including physics-based simulation, image translation, and sim-to-real policy transfer~\cite{narang2021sim,bi2021zeroshot,church2022tactile,su2024sim,yin2023rotating}. More broadly, contact-rich manipulation often requires models or policies that can reason over local
contact dynamics and tactile aliasing~\cite{pang2022global,pmlr-v229-oller23a,zhou2026tactile}. FlexiTac’s matrix-based pressure sensing modality makes tactile simulation and sim-real alignment more tractable. Compared with optical tactile sensors, which produce high-dimensional deformation images that are difficult to simulate faithfully, FlexiTac captures structured contact patterns and pressure distributions mainly along the normal direction. These signals are easier to model, and the sensor also exhibits a stable, approximately linear response region under typical manipulation loads, providing a practical operating regime for calibration and sim-real matching.

For simulation-based learning, FlexiTac can be incorporated into a real-to-sim-to-real pipeline by modeling each tactile pad as a dense set of taxel contact points within a GPU-parallel tactile simulator~\cite{huang2025vtrefine}. Each pad is represented as a dense set of \emph{taxel contact points} uniformly distributed over the surface according to the real sensor resolution (e.g., 12$\times$32), allowing the simulated layout to match the real pad geometry and spatial indexing. At each physics step, we query the signed distance field (SDF) of the contacted object to compute per-taxel penetration depth and relative normal velocity, and then apply a Kelvin--Voigt penalty model (linear spring plus viscous damper) to generate tactile signals. The simulator outputs per-taxel \emph{penetration depth} and \emph{normal force} channels, while \emph{shear forces are omitted} to reduce modeling complexity and improve sim--real consistency.

A key step for sim--real alignment is \emph{tactile calibration}. To match simulated tactile readings to real hardware, we tune only the normal stiffness $k_n$ and damping $k_d$ of the penalty model. Calibration follows a simple procedure: select corresponding taxels in both domains, measure a force--response curve on the real pad, replay the same normal loads in simulation, and iteratively adjust $(k_n, k_d)$ to minimize the curve mismatch; the resulting parameters are then applied across all taxels. To further align distributions, both real and simulated signals use the same normalization rule (including a small noise-floor threshold), and after calibration the normalized tactile histograms closely overlap, which measurably improves downstream fine-tuning stability and reduces sim-to-real degradation.


\vspace{-5pt}
\section{Conclusion}
\vspace{-5pt}
\label{sec:conclusion}

FlexiTac provides a practical, low-cost, and open-source tactile sensing platform that is easy to fabricate, integrate, and scale across diverse robotic systems. By combining manufacturable sensor hardware, lightweight readout electronics, and support for calibration and simulation, FlexiTac lowers the barrier to tactile sensing in robot learning research. Its compatibility with visuo-tactile fusion, cross-embodiment transfer, and simulation-based fine-tuning makes it a useful platform for advancing contact-rich manipulation.

\vspace{-5pt}
\section{Acknowledgments}
\vspace{-5pt}

We thank Yiyue Luo, Devin Murphy, Michael Foshey, Xinyue Zhu, Yunxi Zhu, Naian Tao, Wesley Maa, Jimmy Wang, Binglin Wang, Irving Fang, Hongyu Li, Yixuan Wang, Hanxiao Jiang, Kaifeng Zhang, Changyi Lin, Xuhui Kang, Yuhao Zhou, Pokuang Zhou, Siyu Ma, Chang Yu, Yuqi Liang, Haonan Chen, Rao Fu, Jinzhou Li, Yuming Gu, Fan Cheng, Xinyi Yang, Mingtong Zhang, Baoyu Li, Toru Lin, Haozhi Qi, Iretiayo Akinola, Jie Xu, and Yu-Wei Chao for their valuable discussions.



\bibliography{main}  
\clearpage

\appendix

\end{document}